\documentclass[10pt,twocolumn,letterpaper]{article}

\usepackage{iccv}
\usepackage{times}
\usepackage{epsfig}
\usepackage{graphicx}
\usepackage{amsmath}
\usepackage{amssymb}
\usepackage{bm}
\usepackage{breqn}
\usepackage{subcaption}
\usepackage{booktabs}
\usepackage{multirow}

\usepackage[breaklinks=true,bookmarks=false]{hyperref}

\iccvfinalcopy %

\def\iccvPaperID{****} %

\begin{document}

\title{How You Move Your Head Tells What You Do: Self-supervised Video Representation Learning with Egocentric Cameras and IMU Sensors}

\author{Satoshi Tsutsui\thanks{Research conducted during an internship at Facebook Reality Labs.}\\
Indiana University\\
{\tt\small stsutsui@indiana.edu}
\and
Ruta Desai\\
Facebook Reality Labs\\
{\tt\small rutadesai@fb.com}
\and
Karl Ridgeway\\
Facebook Reality Labs\\
{\tt\small karl.ridgeway@fb.com}
}

\maketitle
\ificcvfinal\thispagestyle{empty}\fi

\begin{abstract}
Understanding users' activities from head-mounted cameras is a fundamental task for Augmented and Virtual Reality (AR/VR) applications. A typical approach is to train a classifier in a supervised manner using data labeled by humans. This approach has limitations due to the expensive annotation cost and the closed coverage of activity labels. A potential way to address these limitations is to use self-supervised learning (SSL). Instead of relying on human annotations, SSL leverages intrinsic properties of data to learn representations. We are particularly interested in learning egocentric video representations benefiting from the head-motion generated by users' daily activities, which can be easily obtained from IMU sensors embedded in AR/VR devices. Towards this goal, we propose a simple but effective approach to learn video representation by learning to tell the corresponding pairs of video clip and head-motion. We demonstrate the effectiveness of our learned representation for recognizing egocentric activities of people and dogs. 
\end{abstract}

\begin{figure}[htb!]
	\includegraphics[width=0.8\columnwidth]{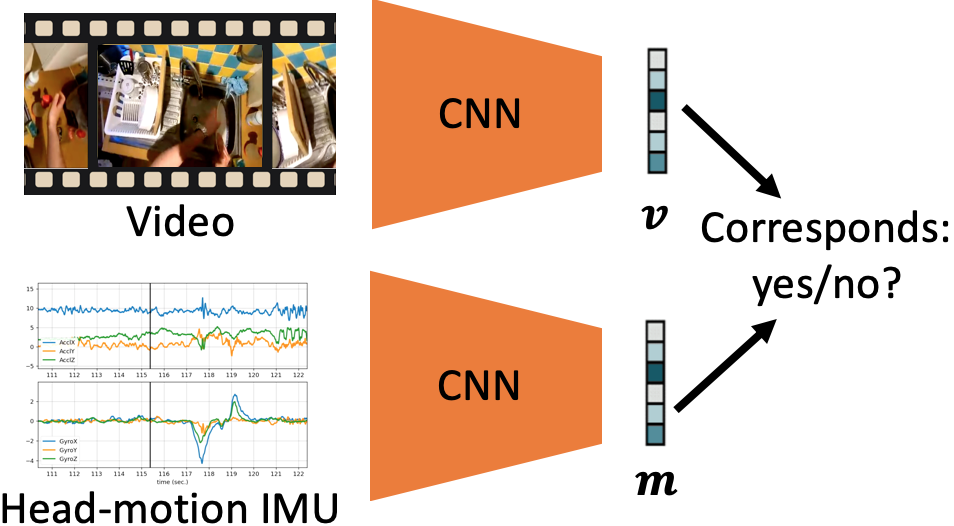}
	\centering
	\caption{Our SSL model maximizes agreement between the representations of video and head-motion captured by a head-mounted camera with IMU sensors. We extract CNN features of a video clip and a head-motion clip, and maximize the similarity between these features, if they are the corresponding pair.  }\label{fig:model}
\end{figure}

\begin{figure*}
    \begin{subfigure}{.26\textwidth}
      \centering
      \includegraphics[keepaspectratio, width=\textwidth]
      {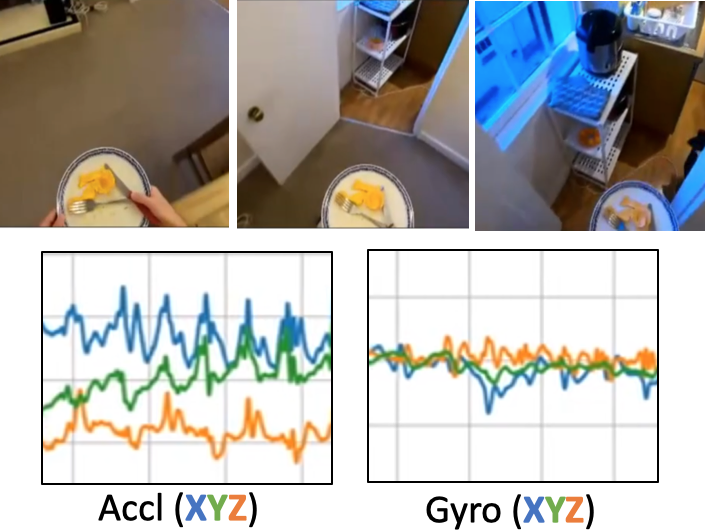}
      \vspace{-4mm}
      \subcaption{Walk/Move}\label{}
    \end{subfigure}\hfill
    \begin{subfigure}{.26\textwidth}
      \centering
      \includegraphics[keepaspectratio, width=\textwidth]
      {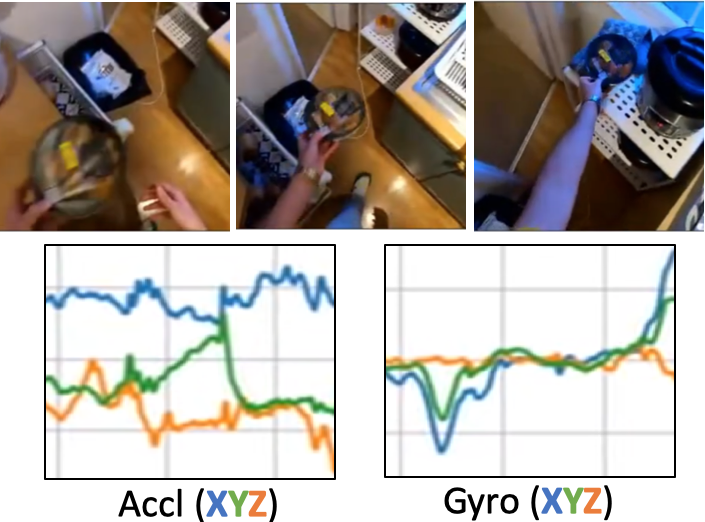}
      \vspace{-4mm}
      \subcaption{Putting an object down}\label{}
    \end{subfigure}\hfill
    \begin{subfigure}{.26\textwidth}
    \centering
      \includegraphics[keepaspectratio, width=\textwidth]
      {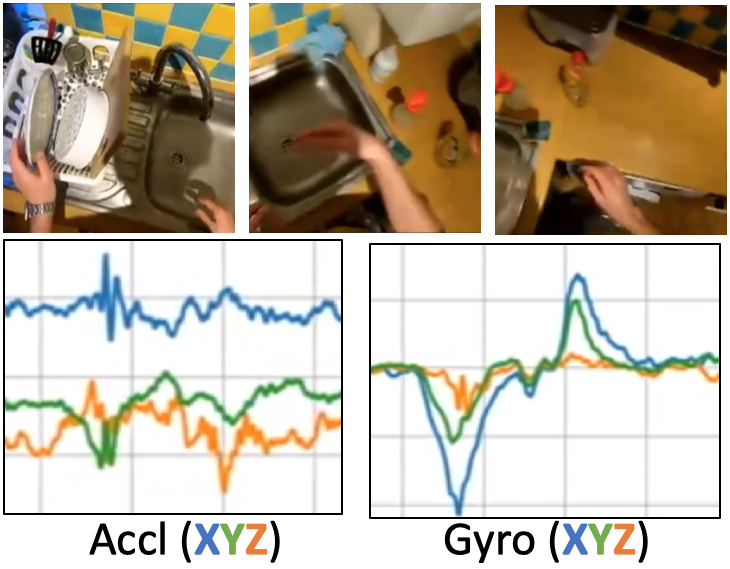}
      \vspace{-4mm}
      \subcaption{Change views/focuses}\label{}
\end{subfigure}
\vspace{-2mm}
 \caption{Head-motion captured synchronously with egocentric videos by the IMU sensors embedded in a head-mounted camera. Head-motion can capture a lot of information about the camera wearer's actions such as walking/moving (a), moving, reaching to an object, putting an object down (b),  or changing view/focus (c). }\label{fig:headmotion}

\end{figure*}

\section{Introduction}
AR/VR technology is finally starting to flourish with the advent of consumer head-mounted devices like Oculus, Ray-Ban Stories, HoloLens, etc. These devices have the potential to fundamentally revolutionize our daily lives and our society -- in a manner similar to what smartphones did in the previous decades. To enable such AR/VR applications, one of the fundamental challenges that requires to be solved is egocentric action recognition -- machine understanding of users' activities from head-mounted cameras.

With the progress of modern computer vision technologies, the now-familiar approach for action recognition is to train convolutional neural networks (CNNs) in a supervised manner using millions of video clips that are manually categorized into egocentric actions. This approach, however, has at least two limitations. First, annotating large enough video clips to train CNNs is very expensive. Second, even if we had unlimited budgets, we would not be able to cover all of the actions that humans could do. 

One of the promising ways to address these limitations is to train CNNs using self-supervised learning (SSL)~\cite{arandjelovic1712objects,chen2020simple}, which has been making rapid progress these days. Instead of relying on human annotations, SSL utilizes intrinsic properties existing in the data (e.g., invariance over data-augmentation~\cite{chen2020simple}, multi-modality of the data~\cite{arandjelovic1712objects}, etc.) for training representations for various downstream tasks including recognition. Inspired by these, we are particularly interested in using head-motion data as a self-supervision signal for egocentric action recognition. Our intuition to leverage head-motion for SSL of egocentric video representations is based on two main factors. First, head-motion is inherently related to users' activities. For instance, head and gaze usually precede picking up/putting down actions; similarly head-motion can also give away movements and change of focus (see Figure~\ref{fig:headmotion}). Secondly, head-motion data can be easily accessible in AR/VR application scenarios via affordable, on-board IMU sensors on egocentric devices. 

To harness the potential of head-motion data for SSL of egocentric video representations, a few fundamental questions need to be answered -- Does head-motion data have unique information that is not captured by the egocentric video representation? If so, what is an effective way to utilize the useful signals in head-motion to benefit egocentric video representation learning? Finally, do the learned representations work better than those trained with SSL on video only data? In this work, we systematically answer these research questions. We empirically show that head-motion can provide additional advantages over video even for fully supervised learning. We then design a simple but effective SSL approach to learn egocentric video representations by classifying pairs of videos and head-motion data based on their correspondence (Figure~\ref{fig:model}). We train our model on the EPIC-KITCHENS dataset using this approach and show the effectiveness of resultant representation for the downstream task of classifying actions in kitchen tasks. Furthermore, we also leverage the same representation to recognize dog-centric activities induced by dogs' head motion, demonstrating that our learned representation generalizes beyond the training domain.

\section{Method}
\paragraph{SSL task formulation}
Inspired by the limitations of labeled datasets, we aim to learn egocentric video representations using SSL for AR/VR applications. In particular, we want to leverage the multimodal data available in AR/VR -- egocentric video and head-motion captured by a head-mounted camera with IMU sensors. SSL usually utilizes a proxy task to train representations without human annotations. For instance, we can learn image representations with a contrastive loss by maximizing the agreement between two different augmented views of the same image~\cite{chen2020simple}. That is, given a pair of randomly augmented images, their representations are encouraged to be similar if they are from the same image, and not if from different images. An extension of this for a multimodal case is to train on the correspondence between two modalities such as audio and video~\cite{arandjelovic1712objects}. Inspired by this audio-visual SSL framework, we propose a binary classification task to match the correspondence between egocentric video and IMU signals of head-motion captured by a head-mounted camera for learning egocentric video representations in AR/VR. 

\paragraph{SSL loss}
To train representations using the above SSL task, we randomly sample a batch of short (2 seconds in our experiments) video clips synchronized with head-motion signals captured by head-mounted IMU sensors. We then extract the feature vectors of video and IMU, compute the pairwise similarities, and encourage the similarities to be high only if they are from the same clip (Figure~\ref{fig:model}). Specifically,  given $\{(\bm{v}_i,\bm{m}_i)\}_{i=1}^N$ -- a batch with $N$ pairs of video and head-motion feature vectors from CNNs, we minimize the following contrastive loss function $L$. 

\begin{dmath}
    L = \frac{1}{N^2} \sum_{i=1}^{N} \sum_{j=1}^{N} \left( 
    \frac{\exp(\mathrm{sim}(\bm v_i, \bm m_j))}{\sum_{k=1}^{N}\exp(\mathrm{sim}(\bm v_i, \bm m_k))}
    \\+  
    \frac{\exp(\mathrm{sim}(\bm v_i, \bm m_j))}{\sum_{k=1}^{N}\exp(\mathrm{sim}(\bm v_k, \bm m_j))}
    \right), 
    \label{eq:loss}
\end{dmath}
where $\mathrm{sim}()$ is the cosine similarity \ie $\mathrm{sim}(\bm v_i, \bm m_j) = {{\bm v_i}^\top {\bm m_j} \over \|{\bm v_i}\| \|{\bm m_j}\|}$.  

After the SSL training, we can use the video representation $\bm{v}_i$ (and also head-motion representation $\bm{m}_i$ if needed) for downstream tasks such as action recognition.

\section{Experiments}

\subsection{Dataset and backbone details}
We use the EPIC-KITCHENS~\cite{Damen2021RESCALING} dataset for all of experiments except the last one. For the final experiment, we use the DogCentric Activity dataset~\cite{iwashita2014first} in order to show generalization of our approach beyond the training dataset. For the EPIC-KITCHENS dataset, we select the video clips accompanied by the corresponding IMU signals of head (camera) motion, and made our own data split of train:validation:test = 30044:3032:4379 based on video ids with no overlapping subjects among the splits. This split has 65 unique test verbs, which means the random guess baseline can achieve the accuracy of 1.5\%. However, due to the biased distribution of actions, the major action (\textit{take}) dominates the 27\% of testing set. For experiments using the DogCentric Activity dataset, we choose the activity categories related to actions with head-motion: \textit{Walk, Shake, Look at left}, and \textit{Look at right}. These four actions are almost balanced and the majority class of \textit{Walk} dominates the 30\% of the dataset. This dataset is small (total 216 video clips and 86 after our selection), so we split into the half-and-half based on dog ids and perform 2-fold cross validation and report the mean accuracy. 

In order to train representations with SSL loss described in eq.~\ref{eq:loss}, we use SlowFast50 as the backbone CNN representation for video and VGG16 for representing the head-motion IMU signals. The spatiotemporal input size of the video CNN is $256 \times 256 \times 48$, corresponding to the width, height, and frame size (with the frame rate of 24fps), respectively. A raw IMU clip is represented with a matrix shaped with $396 \times 6$, corresponding to time (with the frequency of 198Hz) and channels (XYZs for accelerometers and gyroscopes), respectively. Our ways of handling IMU signals are based on \cite{laput2019sensing}, where ordinary image classification CNNs can be used after extracting spectrograms with $n_{fft}=64$. 

\begin{table}[t!]
    \centering
    \begin{tabular}{lccc}
    \toprule
    \multirow{2}{*}{\begin{tabular}[c]{@{}l@{}}Action\\ (Verb)\end{tabular}} & \multicolumn{3}{c}{Model}      \\
    \cmidrule{2-4}
                            & Video & Head-Motion & Ensemble \\
    \midrule 
    Take                    & 74.73 & 52.75       & 76.92    \\
    Put                     & 63.89 & 37.71       & 66.88    \\
    Wash                    & 68.35 & 41.73       & 70.36    \\
    Open                    & 69.35 & 9.58        & 65.90    \\
    Close                   & 47.21 & 18.78       & 42.13   \\
    \bottomrule
    \end{tabular}
    \caption{\emph{Head-motion helps action recognition}: Fully-supervised action classification accuracy (\%) from the video only model,  the head-motion only, and their ensemble.}
    \label{tbl:imuclassificationacc}
\end{table}

\begin{table}[t!]
    \centering
    \begin{tabular}{lcccc}
    \toprule
    \multirow{3}{*}{\begin{tabular}[c]{@{}l@{}}Action\\ (Verb)\end{tabular}} &
      \multicolumn{4}{c}{Correctly Classified by} \\ \cline{2-5} 
     &
      \multicolumn{1}{c}{\begin{tabular}[c]{@{}c@{}} Video\\Only \end{tabular}} &
      \multicolumn{1}{c}{\begin{tabular}[c]{@{}c@{}} Head-Motion\\Only \end{tabular}} &
      \multicolumn{1}{c}{Both} &
      \multicolumn{1}{c}{Neither} \\
      \midrule
        Take  & 384 & 124 & 500 & 175 \\
        Put   & 365 & 120 & 233 & 218 \\
        Wash  & 191 & 59 & 148 & 98 \\
        Open  & 161 & 5 & 20 & 75 \\
        Close & 73 & 17 & 20 & 87 \\
     \bottomrule
    \end{tabular}
\caption{\emph{Head-motion helps action recognition}: The number of correctly classified action clips from video only, head-motion only, both video and head-motion, and neither of them. Even thought video is usually superior due to the richer content, there exits action clips that can be classified using head-motion only.}
\label{tbl:imuclassificationcount}
\end{table}

\subsection{Evaluating the utility of head-motion signals}
Our goal is to utilize head-motion to learn better egocentric video representation for action recognition. However, since video is a rich modality with high fidelity of information, is there any room left for the head motion signals to improve the video representation for action recognition? To answer this question, we perform two preliminary experiments. 

The first experiment is to train an action classifier from head-motion signals and compare with a video only classifier. We expect that the video-based classifier will achieve the higher action classification accuracy. However, if some classes can be correctly classified only by head-motion signals, then this would imply that head-motion indeed has an advantage over video at least for certain classes. We show the classification results of the top five frequent actions (verbs) in Table~\ref{tbl:imuclassificationacc} and \ref{tbl:imuclassificationcount}. The classifier from video has higher accuracy on average, which is what we expected. However, some action clips are correctly classified only from head-motion (Table~\ref{tbl:imuclassificationcount}). Moreover, we also add a simple ensemble model by averaging the probability vectors (\ie outputs after the softmax function) of the two classifiers, and confirm the improvement on the overall accuracy (Table~\ref{tbl:imuclassificationacc}). These results demonstrate an advantage of the head-motion signals over the video. 

\begin{table}[t]
    \centering
    \begin{tabular}{lcc}
    \toprule
    \multirow{2}{*}{} & \multicolumn{2}{c}{Video CNN Pretrained on} \\ \cline{2-3} 
                      & Kinetics           & EPIC-KITCHENS          \\
    \midrule
    Freeze Video CNN  & 87.33              & 86.44                  \\
    Update Video CNN  & \textbf{92.8}               & \textbf{94.73}   \\            \bottomrule
    \end{tabular}
    \caption{\emph{Head-motion information is not inherently captured in existing video representations}: ROC-AUC (\%) on the SSL correspondence task of matching egocentric video and head-motion. Irrespective of how the video representation is pretrained, updating video CNN will give some gain on the task, which suggests a room for improving the video representation by incorporating head-motion. Note that the head-motion CNN is always updated.}
    \label{tbl:sanitycheck}
\end{table}

The second experiment is to see if existing video \textit{representations} (\eg CNN features pretrained on Kinetics) already capture head-motion information or not. This question is important because, if video representations pretrained without head-motion already contain all the information that can be extracted from head-motion, we cannot add any additional value into the video representation by using head-motion. To answer this question, we initialize the video CNN using pretrained weights from Kinetics or EPIC-KITCHENS, and compare the accuracy of our SSL task of matching the correspondence between video and head-motion in two different settings. In the first setting, we train our model (Figure~\ref{fig:model}) with frozen pretreind video CNN and only update the head-motion CNN weights. In the second setting, we update both the video and head-motion CNN weights. We compare the resulting ROC-AUC accuracies of SSL correspondence classification task for both the settings -- without and with the update of the video CNN weights (Table~\ref{tbl:sanitycheck}). We see an increased performance for both CNNs pretrained on Kinetics and EPIC-KITCHENS. Our interpretation is that updating the video CNN weights will not provide any gain on the accuracy if the head-motion information is already embedded in the pretrained video representation. The increased performance indicates that we still have room to improve the video representation by utilizing head-motion. Note that we use ROC-AUC instead of the plain accuracy because most of the pairs are negative correspondence (\ie always classifying as negative achieves the high plain accuracy). 
\begin{table}[t]
    \centering
    \begin{tabular}{lc}
    \toprule
    Representation                    & Acc.         \\
    \midrule
    Fully-Supervised on Kinetics      & 36.58        \\
    Self-Supervised on EPIC-KITCHENS  & 41.94 (Ours) \\
    Fully-Supervised on EPIC-KITCHENS & \textbf{55.61}       \\
    \bottomrule
    \end{tabular}
    \caption{\emph{Head-motion-based SSL pretraining improves downstream classification accuracy}: Action classification accuracy (\%) on EPIC-KITCHENS dataset using representations learned in different ways.}
    \label{tbl:ssldownstream}
\end{table}

\subsection{Leveraging head-motion for action recognition}
\subsubsection{EPIC-KITCHEN action classification using SSL pretrained representation}
After training our model (Figure~\ref{fig:model}) using our SSL task (eq.~\ref{eq:loss}), we can leverage the learned video CNN as a generic video representation backbone for downstream tasks such as egocentric action classification. To test the effectiveness of our video representations learned using SSL, we trained a linear classifier of multiclass logistic regression (or softmax regression) on top of the learned video representation. We also train the same linear classifier on top of the representations learned using fully supervised training for action classification of Kinetics and EPIC-KITCHENS and compare the results (Table~\ref{tbl:ssldownstream}). The classifier that uses the representation learned using SSL achieves the accuracy of 41.94\%. This is higher than the accuracy (27.01\%) of Kinetics representation, and lower than the fully supervised counterpart (55.61\%) of EPIC-KITCHENS. While our SSL representation pre-training is thus effective, it is still behind the fully-supervised counterpart. We therefore wish to close this gap in our future work. 

\begin{table}[t]
    \centering
    \begin{tabular}{lc}
    \toprule
    Representation                    & Acc.         \\
    \midrule
    Fully-Supervised on Kinetics      & 46.98        \\
    Self-Supervised on EPIC-KITCHENS  & \textbf{54.21 (Ours)} \\
    \bottomrule
    \end{tabular}
    \caption{\emph{Video representations pretrained using head-motion-based SSL on kitchen domain are also useful for out of domain tasks}: Action classification accuracy (\%) on DogCentric Activity dataset using a linear classifier on different representations.}
    \label{tbl:ssldog}
\end{table}

\subsubsection{Generalization to DogCentric actions}
We also want to see if the representation learned using our SSL task (eq.~\ref{eq:loss}) generalizes beyond the training domain of kitchens. To test this, we train a linear classifier (multiclass logistic/softmax regression) on DogCentric Activity Dataset using our pretrained SSL representation from EPIC-KICTHEN. We show the results in Table~\ref{tbl:ssldog}. While the classifier based on Kinetics representation has the accuracy of 46.98\%, our SSL representation achieves 54.21\%. This indicates the effectiveness of our SSL approach beyond the training domain. 

\section{Conclusion}
We explored a self-supervised learning (SSL) of video representation by leveraging multimodal egocentric video streams and head-motion captured by IMU sensors for AR/VR applications. While video has much richer information, there’s still room to improve the video representation using head-motion information. Our SSL task, which is tell a simple video-motion correspondence, can train representations effective for egocentric action recognition.

{\small
\bibliographystyle{ieee_fullname}
\bibliography{references}
}

\end{document}